\documentclass[10pt,twocolumn,letterpaper]{article}

\usepackage[pagenumbers]{cvpr} 

\usepackage{algorithm}
\usepackage{algorithmic}

\usepackage{amsfonts}
\usepackage{color}
\usepackage{adjustbox}
\usepackage{booktabs}
\usepackage{multirow}
\usepackage{colortbl}
\usepackage{amssymb}
\usepackage{amsmath}

\usepackage{graphicx}
\usepackage{caption}
\usepackage{lipsum}

\usepackage{fancyhdr}

\def\etal{\textit{et al.~}}
\def\eg{e.g.}
\def\ie{i.e.}
\definecolor{Gray}{gray}{0.9}

\definecolor{cvprblue}{rgb}{0.21,0.49,0.74}
\usepackage[pagebackref,breaklinks,colorlinks,allcolors=cvprblue]{hyperref}

\def\paperID{4588} 
\def\confName{CVPR}
\def\confYear{2026}

\title{GFRRN: Explore the Gaps in Single Image Reflection Removal}

\author{
\begin{tabular}{@{}c@{}}
    Yu Chen\textsuperscript{1,2} \quad 
    Zewei He\textsuperscript{1,2 \textsuperscript{\dag}} \quad 
    Xingyu Liu\textsuperscript{1,2} \quad 
    Zixuan Chen\textsuperscript{3} \quad 
    Zheming Lu\textsuperscript{1,2} \\[2pt]
    \textsuperscript{1}Zhejiang University \quad 
    \textsuperscript{2}Huanjiang Laboratory \quad 
    \textsuperscript{3}The Chinese University of Hong Kong \\[2pt]
\end{tabular}
}

\begin{document}

\twocolumn[{
	\maketitle
	\begin{center}
		\captionsetup{type=figure}
		\includegraphics[width=0.99\textwidth]{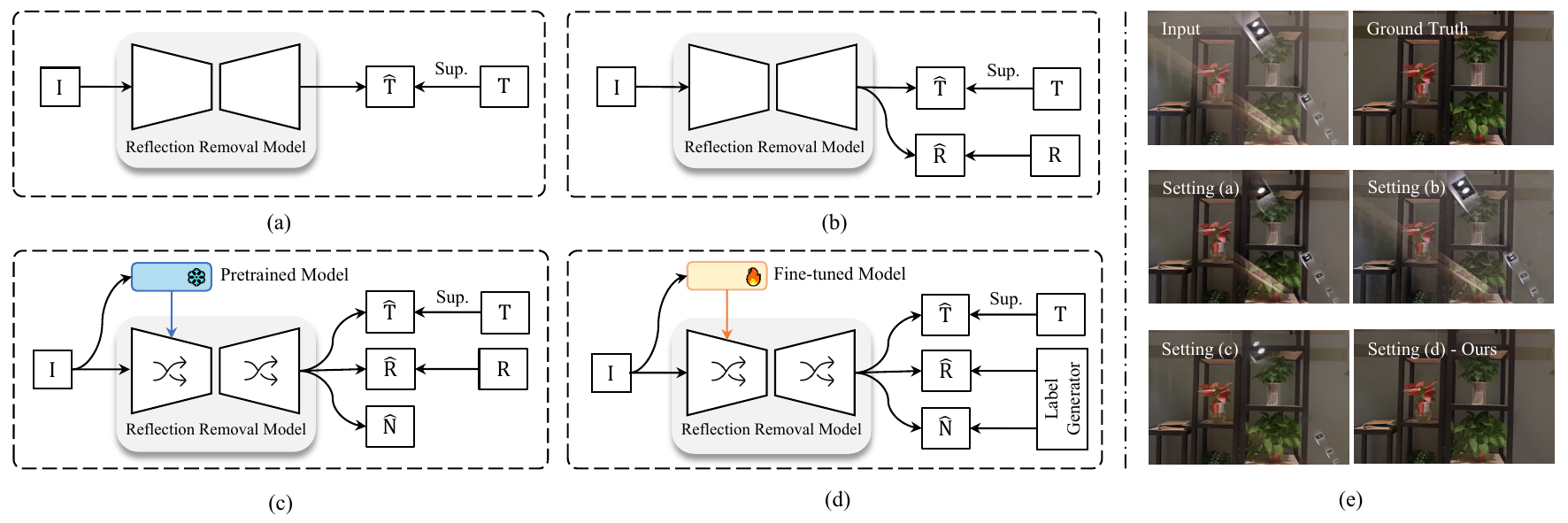}
		\captionof{figure}{Setting (a): Single-stream method like RRW \cite{Zhu2024CVPR-RRW}; Setting (b): Dual-stream method like IBCLN \cite{Li2020CVPR-IBCLN}; Setting (c): Dual-stream method with feature interaction mechanism like DSIT \cite{Hu2024NeurIPS-DSIT}; Setting (d): Ours; (e) Experimental results with different settings.}
		\label{fig1-teaser}
	\end{center}
}]

\renewcommand{\thefootnote}{\fnsymbol{footnote}}
\footnotetext[2]{ Corresponding author: Zewei He (zeweihe@zju.edu.cn).}
\renewcommand{\thefootnote}{\arabic{footnote}}

\begin{abstract}
Prior dual-stream methods with the feature interaction mechanism have achieved remarkable performance in single image reflection removal (SIRR). However, they often struggle with (1) semantic understanding gap between the features of pre-trained models and those of reflection removal models, and (2) reflection label inconsistencies between synthetic and real-world training data. In this work, we first adopt the parameter efficient fine-tuning (PEFT) strategy by integrating several learnable Mona layers into the pre-trained model to align the training directions. Then, a label generator is designed to unify the reflection labels for both synthetic and real-world data. In addition, a Gaussian-based Adaptive Frequency Learning Block (G-AFLB) is proposed to adaptively learn and fuse the frequency priors, and a Dynamic Agent Attention (DAA) is employed as an alternative to window-based attention by dynamically modeling the significance levels across windows (inter-) and within an individual window (intra-). These components constitute our proposed Gap-Free Reflection Removal Network (GFRRN). Extensive experiments demonstrate the effectiveness of our GFRRN, achieving superior performance against state-of-the-art SIRR methods.
\end{abstract}

\section{Introduction}
\label{sec:intro}
When capturing images through glasses or other reflective mediums, reflection artifacts are quite common and manifest as a hybrid mixture of reflected and transmitted components.
Undesired reflections will significantly degrade the imaging quality of the target scene (\ie, transmission layer), further impeding downstream tasks such as object detection and segmentation \cite{Wan2020CVPR,Wieschollek2018ECCV}.
Consequently, the development of robust techniques for transmission-reflection decomposition is highly desired.

The task of single image reflection removal (SIRR) is a long-standing challenge within the realm of blind source separation, primarily attributed to the ill-posedness inherent in the process of decoupling two natural image signals. 
Generally, the observed superimposed image $\mathbf{I}$ can be formulated as follows \cite{Hu2023ICCV-DSRNet}:
\begin{equation}
	\mathbf{I} = \mathbf{T} + \mathbf{R} + \Phi(\mathbf{T},\mathbf{R}),
	\label{eqn1-model}
\end{equation}
where $\mathbf{T}$ and $\mathbf{R}$ denote the target transmission and reflection layers, respectively.
$\Phi(\mathbf{T},\mathbf{R})$ is a residual term \cite{Hu2023ICCV-DSRNet}.
Note that $\Phi$ can represent a group of functions, encompassing a variety of models.

Based on the behavior model in Eqn.~\ref{eqn1-model}, various SIRR methods have been developed.
For example, some single-stream methods \cite{Zhu2024CVPR-RRW,Fan2017ICCV-CEILNet} treat the reflection layer as a form of noise/degradation and focus solely on restoring the transmission layer (refer to Fig.~\ref{fig1-teaser}~(a)).
Some other approaches \cite{Li2020CVPR-IBCLN,Yang2018ECCV-BDN} attempt to simultaneously reconstruct both the reflection layer and the transmission layer (refer to Fig.~\ref{fig1-teaser}~(b)). According to the definition in \cite{Hu2024NeurIPS-DSIT}, we classify them as dual-stream methods.
Currently, dual-stream scheme is a rising trend for SIRR task.
Later, researchers utilized the dual-stream feature interaction mechanism to enhance the information flow \cite{Hu2021NeurIPS-YTMT,Hu2023ICCV-DSRNet,Hu2024NeurIPS-DSIT}.
Among these methods, a prevalent practice is to leverage a pre-trained model for providing semantic information (refer to Fig.~\ref{fig1-teaser}~(c)).

Though existing dual-stream methods (with feature interaction mechanism) \cite{Hu2021NeurIPS-YTMT,Hu2023ICCV-DSRNet,Hu2024NeurIPS-DSIT} have achieved promising results, they ignore some gaps in single image reflection removal (SIRR) task.
(1) \textbf{Semantic gap}: 
Semantic understanding of the input image is helpful in SIRR \cite{Hariharan2015CVPR-Hypercolumns,Zhang2018CVPR,Hu2021NeurIPS-YTMT,Hu2023ICCV-DSRNet,Hu2024NeurIPS-DSIT}.
Typically, such semantic features are extracted by a pre-trained model (VGG \cite{Hu2021NeurIPS-YTMT,Hu2023ICCV-DSRNet} or Swin-Transformer \cite{Hu2024NeurIPS-DSIT}), which does not participate in the gradient back-propagation.
We argue that there exists a semantic gap between the features of pre-trained models and those of reflection removal models.
Aligning the training directions of the pre-trained and the reflection removal models can bridge this gap to some extent, thereby boosting the performance of SIRR task.
(2) \textbf{Training data gap}: 
The training of SIRR models typically involves the concurrent use of both synthetic and real-world data.
However, the labels of the estimated reflection layers for these two types of data are often inconsistent (synthetic: $\mathbf{R}$, real-world: $\mathbf{I}-\mathbf{T}$), which constitutes the so-called ``training data gap" and exerts an adverse impact on the training.
If this gap can be bridged at the data level, it could serve as a general solution, benefiting numerous SIRR approaches.

To remedy the deficiencies discussed above, we, \textbf{for the first time}, introduce parameter efficient fine-tuning technique \cite{Yin2025CVPR-Mona} to adapt the semantic information from the pre-trained model.
This strategy avoids the optimization challenges associated with full fine-tuning (FFT), while effectively achieving alignment between the pre-trained network and the reflection removal network.
Then, we unify the labels of reflection layers for both synthetic and real-world data during the training phase according to our observations and experiments.
Specifically, the low-frequency part of $\mathbf{I}-\mathbf{T}$ is adopted to pose supervision.
Note that, this technique can be applied to existing SIRR models (\eg, DSIT \cite{Hu2024NeurIPS-DSIT}, DSRNet \cite{Hu2023ICCV-DSRNet}), demonstrating its versatility.
In addition, a Gaussian-based adaptive frequency learning block (G-AFLB) is designed to explore and leverage the frequency prior, and a dynamic agent attention (DAA) is employed as an alternative to window-based attention by dynamically modeling the significance levels across windows (inter-window) and within an individual window (intra-window).
At last, by combining the improvements above, we present our \textbf{G}ap-\textbf{F}ree \textbf{R}eflection \textbf{R}emoval \textbf{N}etwork (GFRRN).


\begin{figure*}[t]
	\centering
	\includegraphics[width=0.99\linewidth]{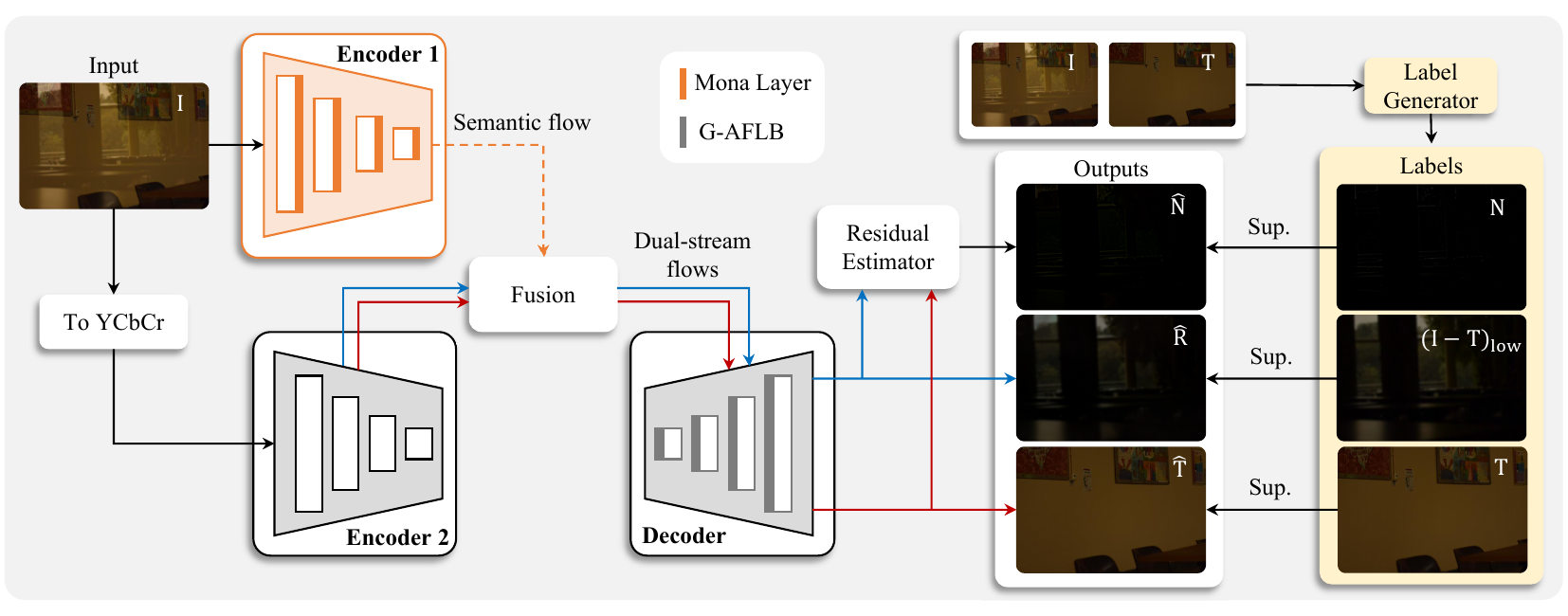}
	\caption{The overall architecture of our GFRRN. It consists of two parallel encoders (\ie, a pre-trained Swin-Transformer with some learnable Mona layers as the \textbf{Encoder 1}, and a dual-stream CNN borrowed from DSIT \cite{Hu2024NeurIPS-DSIT} as the \textbf{Encoder 2}) and a single decoder.}
	\label{fig:fig2}
\end{figure*}

\section{Related work}
\label{sec:relatedwork}
\subsection{Image Reflection Removal}
Currently, multi-frame and multimodel methods \cite{Sarel2004ECCV-separating, Agrawal2005SIGRAPPH-flash, Gai2011TPAMI-blind, sinha2012TOG-rendering, Li2013ICCV-Exploiting, Lei2021CVPR-flash-only, Guo2014CVPR-multiimages, Han2017CVPR-low-rank-matrix, Liu2020CVPR-learning, Sun2016MM-automatic, Xue2015TOG-computational, Yao2025CVPR-PolarFree, Hong2020ICME-infrared} have achieved quite good results. But compared to single-frame methods, their practicality is limited.

Traditional SIRR methods rely on hand-crafted priors based on statistical or physical assumptions. For example, Li \textit{et al.} \cite{Li2014CVPR-smooth} introduced a smoothness prior; Shih \textit{et al.} \cite{Shih2015CVPR-ghosting} utilized ghosting cues; Levin and Weiss \cite{Levin2007PAMI-user} allowed user annotations to guide restoration; and Nikolaos \textit{et al.} \cite{Arvanitopoulos2017CVPR-Suppression} combined a Laplacian data term with gradient sparsity. Although these priors do not always hold in practice, they have significantly advanced SIRR development.

With the rise of deep learning, data-driven approaches have gradually replaced traditional methods and become dominant in SIRR. CEILNet \cite{Fan2017ICCV-CEILNet} explicitly divides the task into two stages: estimate the edge map and restored image. Zhang \etal \cite{Zhang2018CVPR} proposed a convolutional network trained with perceptual and exclusion losses. ERRNet \cite{Wei2019CVPR-ERRNet} provides a way to utilize non-aligned image pairs for training. Although these methods achieve impressive results, they often discard useful information in the reflection layer. IBCLN \cite{Li2020CVPR-IBCLN} uses LSTM units to iteratively recover both reflection and transmission layers. YTMT \cite{Hu2021NeurIPS-YTMT} introduces a dual-stream interaction module based on ReLU and negative ReLU. DSRNet \cite{Hu2023ICCV-DSRNet} proposes a learnable residual mechanism that unifies existing physical models and designs a MUGI block for dual-stream interaction. DSIT \cite{Hu2024NeurIPS-DSIT} employs attention mechanisms to further improve interaction accuracy. Dong \etal \cite{Dong2021ICCV-LASIRR} precisely locate reflection and remove it. RRW \cite{Zhu2024CVPR-RRW} proposes a cascaded strategy network and an efficient data acquisition pipeline. FIRM \cite{Chen2025AAAI-Firm} achieves accurate reflection removal by allowing users to manually specify reflection regions. DExNet \cite{Huang2025TPAMI-DExNet} introduces a lightweight deep unfolding network. RDNet \cite{Zhao2025CVPR-RDNet} designs a reversible network that performs well on SIRR. These methods utilize information from reflection and transmission layers, but they requires ITR-aligned triplet datasets.

\subsection{Parameter Efficient Fine Tuning}
Recent studies have shown that pretrained models from high-level vision tasks can assist low-level restoration tasks. For instance, DSRNet, DSIT in SIRR, or SGGLC-Net \cite{Fan2025TIM-SGGLCNet} in super-resolution. However, there still exist inevitable gaps between high-level and low-level tasks, making fine-tuning of pretrained models essential. The most straightforward approach is Full Fine-Tuning (FFT), but it incurs high computational costs. Parameter-Efficient Fine-Tuning (PEFT) offers an effective solution to these challenges.
The first strategy is to freeze most parameters of the pretrained model and only train a small subset, such as BitFit \cite{Zaken2022ACL-Bitfit} only fine-tunes the bias, Norm Tuning \cite{giannou2023arXiv-Norm-Turning} only fine-tunes the norm layer. The second strategy reparameterizes parts of the pretrained model, LoRA \cite{Hu2022ICLR-Lora} is the most representative example. The third strategy fixes all pretrained parameters while adding small trainable modules, such as AdaptFormer \cite{Chen2022NeurIPS-AdaptFormer}, Mona \cite{Yin2025CVPR-Mona}, VPT \cite{jia2022ECCV-VPT} and P-Tuning \cite{liu2022ACL-PTune}.

\section{Methodology}

\subsection{Basic model}
Considering that the reflection layer may contain valuable information and can provide better regularization for the restoration of the transmission layer, we also choose to employ a dual-stream framework.
Among the existing approaches, the most representative ones include YTYM \cite{Hu2021NeurIPS-YTMT}, DSRNet \cite{Hu2023ICCV-DSRNet}, and DSIT \cite{Hu2024NeurIPS-DSIT}.
They typically incorporate a pre-trained model (\eg, VGG or Swin-Transformer) to provide high-level semantic information, which is then injected into the dual-stream flows via a certain fusion scheme.

\begin{figure}[t]
	\centering
	\includegraphics[width=0.99\linewidth]{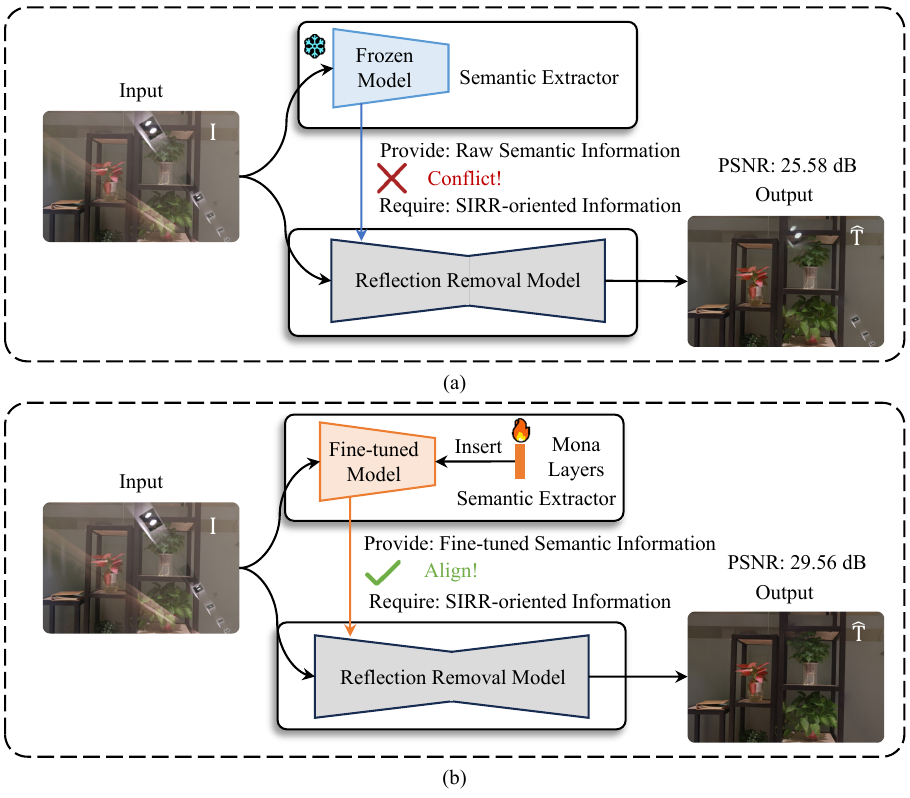}
	\caption{(a) Semantic gap exists between the pre-trained model and the reflection removal model. (b) A cognitive-inspired Mona-tuning technique is proposed to bridge the semantic gap.}
	\label{fig:fig3}
\end{figure}

Following their basic configurations, the overall architecture of our Gap-Free Reflection Removal Network (GFRRN) is illustrated in Fig.~\ref{fig:fig2}, which consists of two parallel encoders (\ie, a pre-trained Swin-Transformer with some learnable Mona layers as the \textbf{Encoder 1} for extracting the global priors, and a task-specific dual-stream CNN borrowed from DSIT \cite{Hu2024NeurIPS-DSIT} as the \textbf{Encoder 2} for extracting the local priors) and a single decoder. 
The estimated reflection $\hat{\mathbf{R}}$ and transmission $\hat{\mathbf{T}}$ are employed to predict the residual term $\hat{\mathbf{N}}$ via a residual estimator, which mainly composed of a NAFBlock \cite{Chen2022ECCV-NAFNet}. 

Our GFRRN has a similar architecture to the DSIT \cite{Hu2024NeurIPS-DSIT}, yet with three main differences: 
(1) An adapter-based tuning strategy is introduced into the SwinBlock \cite{Liu2021ICCV-Swin}, which fixes the pre-trained weights and solely updates the weights in adapters. Specifically, the \textbf{M}ulti-c\textbf{o}g\textbf{n}itive visual \textbf{a}dapter (Mona) \cite{Yin2025CVPR-Mona} is adopted with vision-friendly filters to transfer the pre-trained knowledge through multiple cognitive perspectives.
(2) The supervision label of the estimated reflection is designed as the low-frequency part of $\mathbf{I}-\mathbf{T}$ instead of direct $\mathbf{I}-\mathbf{T}$, to ensure that it exclusively contains information from the reflection layer.
(3) In the decoder part, a Gaussian-based adaptive frequency learning block (G-AFLB) is employed to learn and integrate the frequency priors. Moreover, the window-based multi-head self-attention (W-MSA) is replaced with proposed dynamic agent attention (DAA).
These components will be elaborated in detail in the following subsections.

\subsection{Mona-tuning}
The Swin-Transformer \cite{Liu2021ICCV-Swin} used in our GFRRN is initially designed for the classification task, which primarily emphasize high-level semantic information.
This focus differs from that of the image restoration task (precisely, reflection removal), as the latter places greater importance on low-level texture details for dense prediction.
Therefore, semantic gap exists between the pre-trained model (\ie, Swin-Transformer) and the reflection removal model (see in Fig.~\ref{fig:fig3}~(a)).


While full fine-tuning (FFT) of the Swin-Transformer appears to be a straightforward solution, its performance remains sub-optimal (as show in Table~\ref{tab:tab3}).
This may stem from the model's excessive trainable parameters, which cannot be effectively optimized given our dataset's limited scale compared to ImageNet \cite{Deng2009CVPR-ImageNet}.

Recent delta-tuning or parameter efficient fine-tuning (PEFT) methods \cite{Chen2022NeurIPS-AdaptFormer,Zaken2022ACL-Bitfit,Yin2025CVPR-Mona} provide new options for this problem.
As show in Fig.~\ref{fig:fig3}~(b), we opt for a cognitive-inspired Mona-tuning technique, which integrates convolution-based filters to bridge the semantic gap by transferring visual knowledge from the pre-trained model to the reflection removal task.
Specifically, we insert Mona layers after MSA and MLP in each SwinBlock of Swin-Transformer \cite{Liu2021ICCV-Swin}. \textbf{More details can be found in our supplementary material.}
During the training phase, only the weights of inserted Mona layers are updated.
To the best of our knowledge, this is the first work to apply the PEFT technique to the SIRR task.

\subsection{Unified label}

In-depth analysis demonstrates that the issue of data gap primarily arises from the divergence in supervision labels employed during the training process.
Specifically, when handling synthetic datasets, models are typically supervised using reflection images denoted as $\mathbf{R}$, but switch to using residual images (constructed as $\mathbf{I} - \mathbf{T}$) \footnote{Most real training datasets do not contain reflection images, since the difficulty in acquisition.} as supervision for real datasets \cite{Huang2025Arxiv}. 
This discrepancy in supervision labels between synthetic and real datasets substantially undermines the model's capacity for generalization.

\begin{figure}[t]
	\centering
	\includegraphics[width=0.99\linewidth]{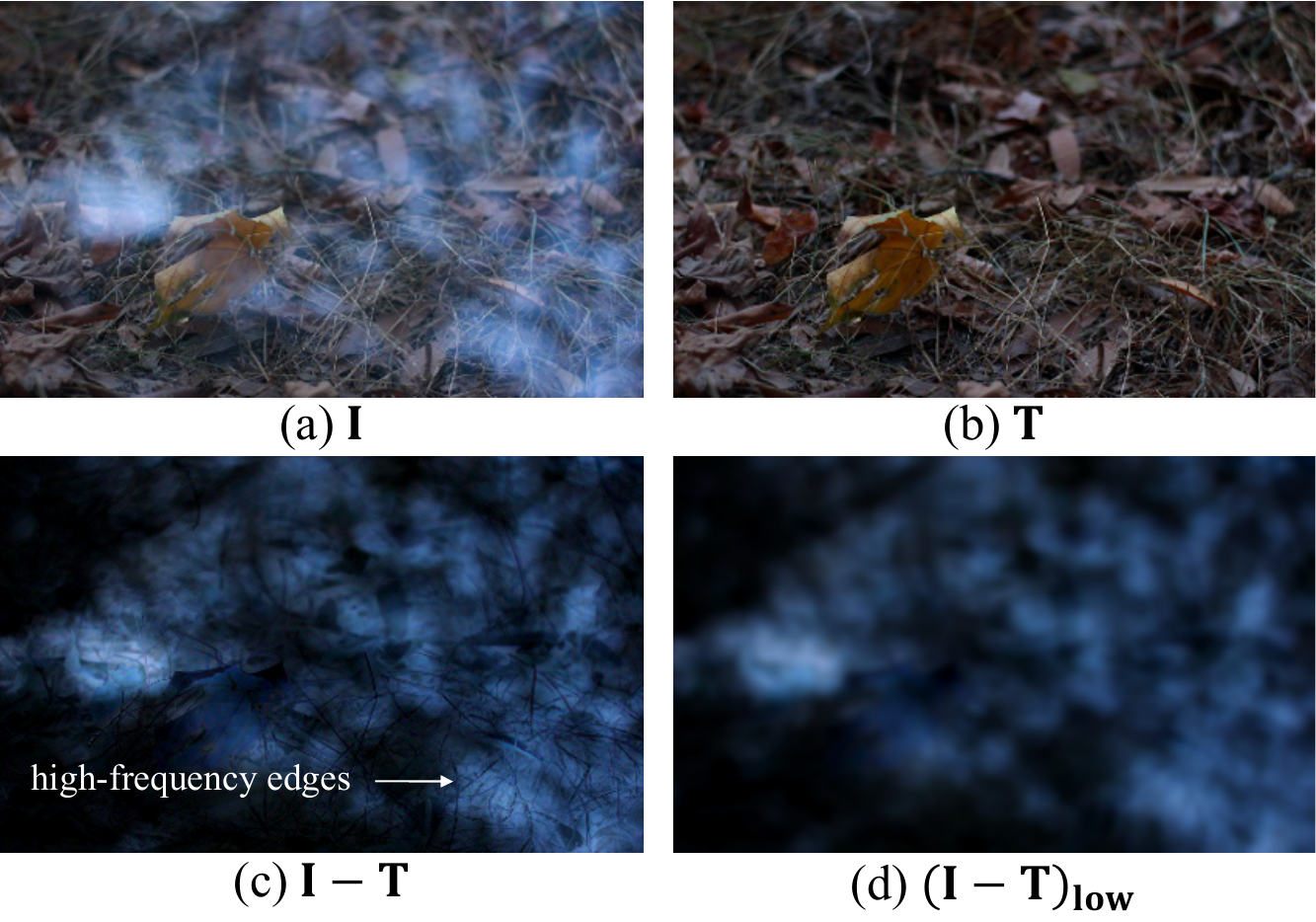}
	\caption{(a) A degraded image $\mathbf{I}$ from Real dataset. (b) The label of corresponding transmission layer. (c) $\mathbf{I} - \mathbf{T}$ for supervising the reflection layer. (d) Our unified label. $(\mathbf{I} - \mathbf{T})_{\text{low}}$ denotes the low-frequency part.}
	\label{fig:fig4}
\end{figure}

We firstly employ $\mathbf{I} - \mathbf{T}$ as the reflection layer label for synthetic data to maintain consistency with real-world data.
However, our experimental results reveal a significant performance drop (as shown in Table~\ref{tab:tab4}).
Upon observation of Fig.~\ref{fig:fig4}~(c), we find that $\mathbf{I} - \mathbf{T}$ contains obvious high-frequency information from the transmission layer (\eg, edges), which may lead the network to mistakenly classify some transmission-related information as part of the learned reflection layer.
According to Eqn.~\ref{eqn1-model}, $\mathbf{I} - \mathbf{T}$ inherently contains information from the residual term, which is consistent with our observation.

We believe that a reasonable reflection layer label should, to the greatest possible extent, avoid incorporating information from the transmission layer.
Based on this, we employ a label generator to filter out the high-frequency edges within $\mathbf{I} - \mathbf{T}$, which is originated from the transmission layer .
In our implementation, the label generator is just a simple 2D low-pass filter, and the reflection label $(\mathbf{I} - \mathbf{T})_{\text{low}}$ is illustrated in Fig.~\ref{fig:fig4}~(d).

Note that, this reflection label is not rigorously accurate.
Some information of the reflection layer may also be filtered out under specific circumstances (into the residual term).
Additionally, we conduct direct supervision on the learned residual term $\hat{\mathbf{N}}$, with the corresponding label denoted as $\mathbf{I} - \mathbf{T} - (\mathbf{I} - \mathbf{T})_{\text{low}}$.
By this way, the filtered-out information will be encapsulated in the learnable residual term, and this operation can provide regularization for the estimation of both the reflection and transmission layers.

\subsection{G-AFLB \& DAA in decoder}

\noindent \textbf{G-AFLB.}
When capturing images, the reflection layer exhibits varying degrees of blurriness depending on its relative position with respect to the depth of field.
Though the SIRR task inherently exhibits typical frequency priors, frequency information is seldom explicitly utilized in the reflection removal models.
Therefore, to fully explore and leverage frequency information, inspired by \cite{Cui2025ICLR-adaIR} we design a Gaussian-based Adaptive Frequency Learning Block (G-AFLB), which has two ingenious considerations.
On one hand, we utilize smoothed Gaussian coefficients to replace the binary frequency boundary for suppressing the Gibbs effect. On the other hand, it can adaptively match the degree of blurriness of the reflection layer.
\textbf{The details of our G-AFLB can be found in our supplementary material.}

\begin{figure}[t]
	\centering
	\includegraphics[width=0.99\linewidth]{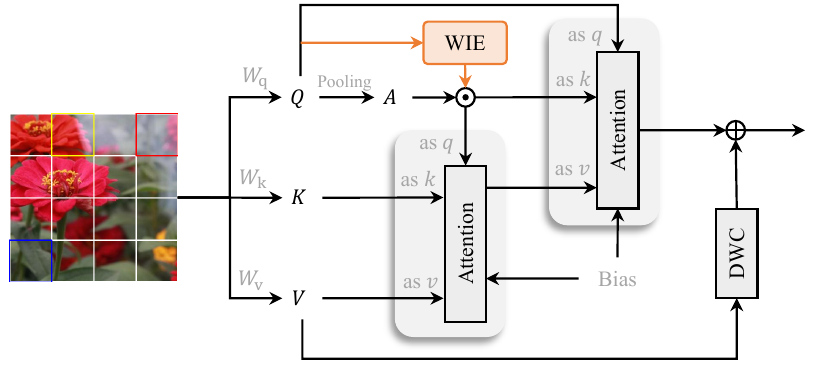}
	\caption{Details of our proposed dynamic agent attention (DAA).}
	\label{fig:fig5}
\end{figure}

\noindent \textbf{DAA.}
Window-based multi-head self-attention (W-MSA) is the core component of the feature interaction mechanism \cite{Hu2024NeurIPS-DSIT}.
We first employ the agent attention \cite{Han2024ECCV-Agent} to replace original W-MSA in consideration of computational efficiency.
However, agent attention neglects the reflection discrepancies across different windows.
As shown in Fig.~\ref{fig:fig5}, the red window is entirely obscured by reflection, whereas the blue one is free of reflection, and the yellow one contains partial.
According to this observation, we design a window-based importance estimator (WIE) in the query branch to assign the self-learned importance weight to each window.
\textbf{The details of WIE can be found in our supplementary material.}
As shown in Fig.~\ref{fig:fig5}, combining the WIE with agent attention makes up our dynamic agent attention (DAA). 

\subsection{Loss function}
Given the observed superimposed image $\mathbf{I}$ and transmission layer label $\mathbf{T}$, we first generate the labels for the reflection layer $\mathbf{R}=(\mathbf{I} - \mathbf{T})_{\text{low}}$ and residual term $\mathbf{N}=\mathbf{I} - \mathbf{T} - (\mathbf{I} - \mathbf{T})_{\text{low}}$.

\noindent \textbf{Content loss.}
We supervise $\hat{\mathbf{T}}$, $\hat{\mathbf{R}}$ in both the spatial and gradient domains. Furthermore, we find regularizing $\hat{\mathbf{N}}$ improves model performance, so we also incorporate it into the content loss.
\begin{equation}
	\begin{split}
		\mathcal{L}_{con} &:=  \|\hat{\mathbf{T}} - \mathbf{T}\|_{2}^{2} + \|\hat{\mathbf{R}} - \mathbf{R}\|_{2}^{2} + \alpha\|\hat{\mathbf{N}} - \mathbf{N}\|_{2}^{2} \\
		& + \beta(\|\nabla \hat{\mathbf{T}} - \nabla \mathbf{T}\|_{1} + \|\nabla \hat{\mathbf{R}} - \nabla \mathbf{R}\|_{1} \\
		& + \alpha\|\nabla \hat{\mathbf{N}} - \nabla \mathbf{N}\|_{1}),
	\end{split}
\end{equation}
where $\|\cdot\|_{2}$ and $\|\cdot\|_{1}$ represent the $\ell_{2}$ and $\ell_{1}$ norms, respectively. $\nabla$ stands for the gradient operator. In our implementation, $\alpha = 0.3$ and $\beta=0.6$.

\noindent \textbf{Exclusion loss.}
We also introduce exclusion loss \cite{Zhang2018CVPR} to ensure the structural independence of $\hat{\mathbf{T}}$, $\hat{\mathbf{R}}$:
\begin{equation}
	\begin{aligned}
		\mathcal{L}_{exc} := & \frac{1}{3} \sum_{n=0}^{2} \| \mathcal{D} \left( \hat{\mathbf{T}}^{\downarrow n}, \hat{\mathbf{R}}^{\downarrow n} \right) \|_{2}^{2},\\
		\mathcal{D}(\hat{\mathbf{T}}, \hat{\mathbf{R}}) := & \tanh \left( \xi_{1} | \nabla \hat{\mathbf{T}} | \right) \circ \tanh \left( \xi_{2} | \nabla \hat{\mathbf{R}} | \right),
	\end{aligned}
\end{equation}
where $\hat{\mathbf{T}}^{\downarrow n}, \hat{\mathbf{R}}^{\downarrow n}$ are $2^{n}$ down-sampled version of $\hat{\mathbf{T}}$ and $\hat{\mathbf{R}}$. $\xi_{1}$ and $\xi_{2}$ are normalization factors.

\noindent \textbf{Perceptual Loss.}
To further ensure the restoration quality of the transmission layer:
\begin{equation}
	\mathcal{L}_{per} := \sum_{i} \omega_{i} \| \phi_{i}(\hat{\mathbf{T}}) - \phi_{i}(\mathbf{T}) \|_{1},
\end{equation}
where $\phi_{i}(\cdot)$ represents the intermediate feature of the pre-trained VGG-19, $i \in \{2, 7, 12, 21, 30\}$ is the layer ID.

\noindent \textbf{Reconstruction Loss.}
We further constrain image recovery quality through reconstruction:
\begin{equation}
	\mathcal{L}_{rec} := \|\mathbf{I} - \hat{\mathbf{T}} - \hat{\mathbf{R}}-\hat{\mathbf{N}}\|_{1}.
\end{equation}

\noindent \textbf{Total Loss.}
The full training objectives $\mathcal{L}_{total}$ is defined as follows:
\begin{equation}
	\mathcal{L}_{total} := \mathcal{L}_{con} + \mathcal{L}_{exc} + \lambda_{1} \mathcal{L}_{per}+\lambda_2\mathcal{L}_{rec},
\end{equation}
where $\lambda_{1} = 0.01$, and $\lambda_{2} = 0.2$.

%
%
%
\section{Experiments}
\label{sec:exp}

\begin{table*}[t]
	\footnotesize
	\caption{Benchmark results of various SIRR methods on Real20, Object200, Postcard199, Wild55, Nature20 testing datasets. \textbf{Bold} and \underline{underlined} indicate the best and the second best performance, respectively. For fair comparison, we use the digits in their original paper or test with the pre-trained model they provided.}
	\label{tab:benchmark}
	\centering
	\begin{tabular}{l|l|cccccccccc|cc}
		\toprule[1.2pt]
		\multirow{2}{*}{Methods} & \multirow{2}{*}{Venues}& \multicolumn{2}{c}{Real20} & \multicolumn{2}{c}{Object200} & \multicolumn{2}{c}{Postcard199} & \multicolumn{2}{c}{Wild55} & \multicolumn{2}{c|}{Nature20} & \multicolumn{2}{c}{Average494} \\
		&& PSNR & SSIM & PSNR & SSIM & PSNR & SSIM & PSNR & SSIM & PSNR & SSIM & PSNR & SSIM \\
		\midrule \midrule
		ERRNet \cite{Wei2019CVPR-ERRNet}     &CVPR'19& 22.89 & 0.803 & 24.87 & 0.896 & 22.04 & 0.876 & 24.25 & 0.853 & 20.58 & 0.756 & 23.41 & 0.874 \\
		IBCLN \cite{Li2020CVPR-IBCLN}      &CVPR'20& 21.86 & 0.762 & 24.87 & 0.893 & 23.39 & 0.875 & 24.71 & 0.886 & 23.57 & 0.786 & 24.08 & 0.875 \\
		LASIRR \cite{Dong2021ICCV-LASIRR}     &ICCV'21& 23.34 & 0.812 & 24.36 & 0.898 & 23.72 & 0.903 & 25.73 & 0.902 & 23.45 & 0.808 & 24.18 & 0.893 \\
		YTMT \cite{Hu2021NeurIPS-YTMT}       &NeurIPS'21& 23.26 & 0.806 & 24.87 & 0.896 & 22.91 & 0.884 & 25.48 & 0.890 & 23.85 & 0.810 & 24.04 & 0.880 \\
		RobustSIRR \cite{Song2023CVPR-RobustSIRR}  &CVPR'23& 23.61 & 0.835 & 24.90 & 0.917 & 19.91 & 0.868 & 23.67 & 0.884 & 20.97 & 0.764 & 22.54 & 0.884 \\
		DSRNet \cite{Hu2023ICCV-DSRNet}      &ICCV'23& 23.91 & 0.818 & 26.74 & 0.920 & 24.83 & 0.911 & 26.11 & 0.906 & 25.22 & 0.832 & 25.72 & 0.907 \\
		DURRNet \cite{Huang2024ICASSP-DURRnet}    &ICASSP'24& 23.80 & 0.810 & 24.52 & 0.891 & 22.26 & 0.866 & 25.62 & 0.899 & 24.24 & 0.812 & 23.69 & 0.875 \\
		RRW \cite{Zhu2024CVPR-RRW}        &CVPR'24& 23.82 & 0.817 & 26.55 & 0.927 & 24.03 & 0.903 & 26.51 & 0.913 & 25.96 & 0.843 & 25.40 & 0.908 \\
		DSIT \cite{Hu2024NeurIPS-DSIT}        &NeurIPS'24& 25.22 & 0.836 & \underline {27.27} & \underline {0.932} & 25.58 & \underline {0.922} & 27.40 & \underline {0.918} & \underline {26.77} & \underline {0.847} & 26.50 & \underline {0.919} \\
		DExNet \cite{Huang2025TPAMI-DExNet}     &TPAMI'25& 23.50 & 0.817 & 26.38 & 0.916 & 25.52 & 0.918 & 26.95 & 0.908 & 24.66 & 0.837 & 25.91 & 0.909 \\
		RDNet \cite{Zhao2025CVPR-RDNet}      &CVPR'25& \underline {25.58} & \underline {0.846} & 26.78 & 0.921 & \underline {26.33} & \underline {0.922} & \underline {27.70} & 0.915 & 26.21 & 0.842 &\underline {26.63} & 0.915 \\
		\midrule
		\rowcolor{Gray}Ours        &-& \textbf{25.84} & \textbf{0.847} & \textbf{27.73} & \textbf{0.936} & \textbf{26.80} & \textbf{0.937} & \textbf{28.29} & \textbf{0.926} & \textbf{27.39} & \textbf{0.861} & \textbf{27.33} & \textbf{0.929} \\
		\bottomrule[1.2pt]
	\end{tabular}
\end{table*}

\begin{figure*}[t]
	\centering
	\includegraphics[width=0.99\linewidth]{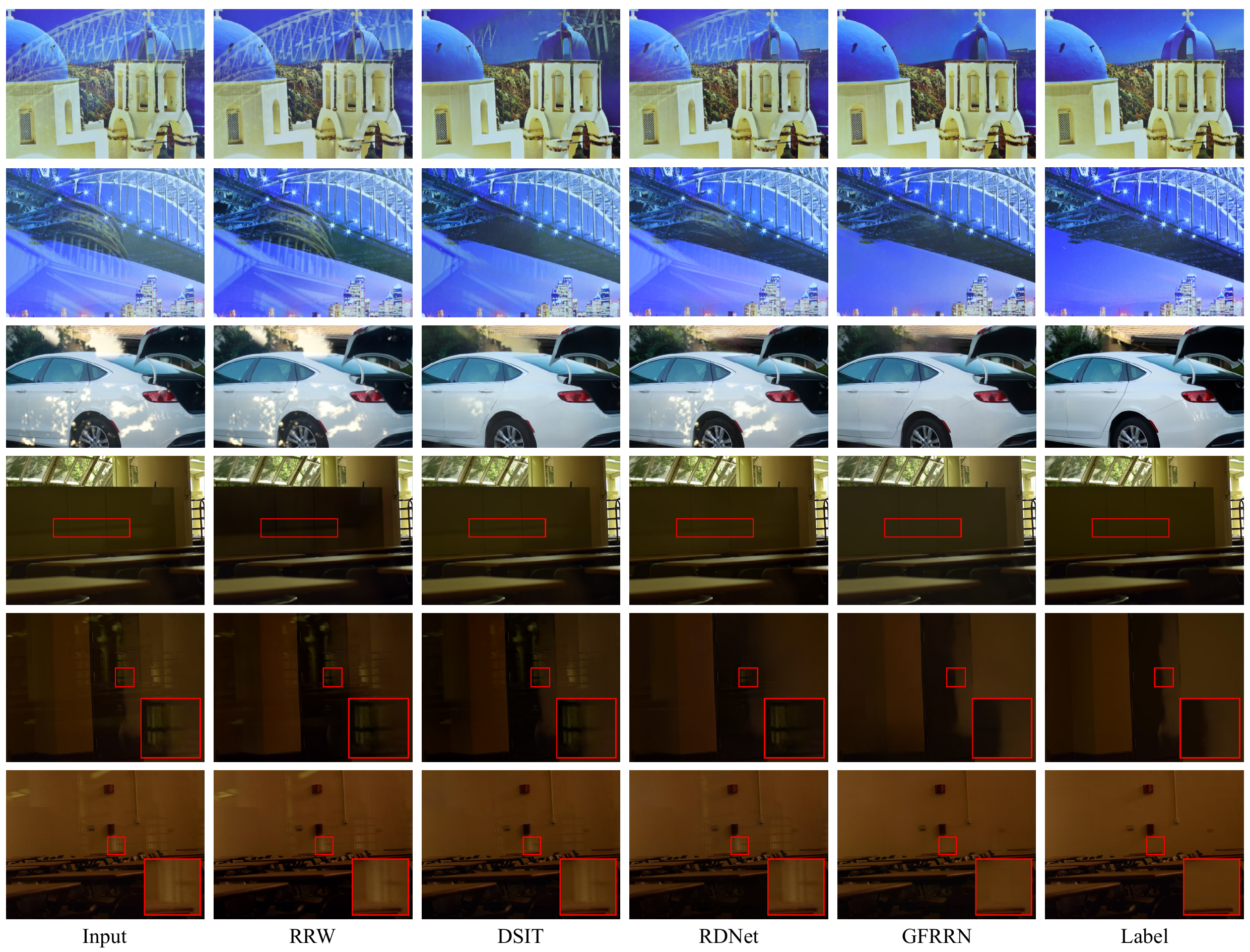}
	\caption{Visual comparisons between our GFRRN and several state-of-the-art methods on samples from SIR$^2$ and Real20 datasets. Please zoom-in on screen for more details.}
	\label{fig:fig6}
\end{figure*}

\subsection{Implementation details}

\noindent \textbf{Datasets.}
Following previous works \cite{Hu2023ICCV-DSRNet,Hu2024NeurIPS-DSIT,Zhao2025CVPR-RDNet}, our training datasets include both synthetic and real-world images.  
We randomly sample 5K synthetic pairs from the PASCAL VOC dataset \cite{Everingham2010IJCV-PascalVOC} in each epoch.
90 image pairs from ``Real'' dataset \cite{Zhang2018CVPR} and 200 image pairs from ``Nature'' dataset \cite{Li2020CVPR-IBCLN} are also included for training.
All the images are resized to $384\times 384$.
For the testing, we employ 5 commonly-used testing datasets including: Real20 \cite{Zhang2018CVPR}, Nature20 \cite{Li2020CVPR-IBCLN}, and three subsets of the SIR$^2$ datasets \cite{Wan2017ICCV-SIR2} (\ie, Object200, Postcard199, Wild55).

\noindent \textbf{Training strategy.} 
During the training phase, the Adam optimizer with default parameters is adopted to train our GFRRN for 60 epochs.
The learning rate is fixed as $1e^{-4}$ with a batch size of 1 on a single NVIDIA RTX A6000 GPU.
Some unmentioned details are aligned with previous DSIT model \cite{Hu2024NeurIPS-DSIT}.

\subsection{Performance evaluation}
\noindent \textbf{Quantitative comparison.}
In this section, we compare our GFRRN with 11 state-of-the-art SIRR methods including: ERRNet \cite{Wei2019CVPR-ERRNet}, IBCLN \cite{Li2020CVPR-IBCLN}, LASIRR \cite{Dong2021ICCV-LASIRR}, YTMT \cite{Hu2021NeurIPS-YTMT}, RobustSIRR \cite{Song2023CVPR-RobustSIRR}, DSRNet \cite{Hu2023ICCV-DSRNet}, DURRNet \cite{Huang2024ICASSP-DURRnet}, RRW \cite{Zhu2024CVPR-RRW}, DSIT \cite{Hu2024NeurIPS-DSIT}, DExNet \cite{Huang2025TPAMI-DExNet}, RDNet \cite{Zhao2025CVPR-RDNet} on 5 real-world testing datasets.
Some of these methods have two different training settings.
We choose the setting matched with ours to report the metrics.
Note that, RRW \cite{Zhu2024CVPR-RRW} employs extra training pairs.

Table~\ref{tab:benchmark} shows the benchmark results on our testing datasets.
We observe that our GFRRN ranks the first in all comparisons, gaining 0.7 dB and 0.01 improvements in terms of average PSNR and SSIM, respectively.
Considering that the five real-world datasets contain a variety of scenes, illumination conditions, and glass thicknesses, achieving optimal performance simultaneously across all metrics on these datasets is quite challenging.

\noindent \textbf{Quantitative comparison.}
Fig.~\ref{fig:fig6} shows the visual results of RRW, DSIT, RDNet and our GFRRN on samples from SIR$^2$ and Real20 datasets.
\textbf{Due to the limited space, additional examples can be found in our supplementary material.}
The samples in the top two rows are sourced from Postcard199, and we can clearly observe that our GFRRN removes the reflection effectively, revealing rich texture and correct color information.
In contrast, other methods struggle to remove the reflection completely.
The samples in the third row is sourced from Real20 with intense specular reflection on the vehicle's surface.
Our GFRRN can handle this kind of reflection successfully.
DSIT \cite{Hu2024NeurIPS-DSIT} also removes the majority, yet a small amount still remains.
The samples in the bottom three rows are sourced from Wild55.
The reflections in these images are relatively weak and concealed within the textures of the transmission layers.
Our GFRRN produces clearer and more visually appealing results.
Overall, our results are more visually favorable and contains fewer residual reflection components, which is consistent with the evaluation metrics.

\subsection{Ablation study}
To validate the effectiveness of our GFRRN, we conduct ablation study on the key components (\ie, Mona-tuning, unified label, G-AFLB, DAA), followed by in-depth explorations of their designs.

\noindent \textbf{Key components.}
We first perform ablation study by removing corresponding component, and the experimental results are shown in Table~\ref{tab:tab2}.
w/o DAA means using W-MSA instead of DAA, w/o G-AFLB means this block is removed, w/o Mona-tuning means the pre-trained model is frozen, w/o unified label means the reflection label used is aligned with previous works \cite{Hu2024NeurIPS-DSIT,Hu2021NeurIPS-YTMT,Zhao2025CVPR-RDNet}.
As we can see, each component plays an indispensable role, and omitting any of them will compromise the model's performance in terms of PSNR and SSIM.

\begin{table}[t]
	\footnotesize
	\caption{Ablation study on different configurations. The PSNR and SSIM values are computed across all five testing datasets and subsequently averaged.}
	\label{tab:tab2}
	\centering
	\begin{tabular}{l|cc}
		\toprule[1.2pt]
		Model & PSNR & SSIM \\
		\midrule\midrule
		w/o DAA & 26.91 & 0.919 \\
		w/o G-AFLB & 27.02 & 0.923 \\
		w/o Mona-tuning & 26.70 & 0.920 \\
		w/o unified label & 26.96 & 0.920  \\
		\midrule
		\rowcolor{Gray}Ours & 27.33 & 0.929 \\
		\bottomrule[1.2pt]
	\end{tabular}
\end{table}

\begin{figure*}[t]
	\centering
	\includegraphics[width=0.99\linewidth]{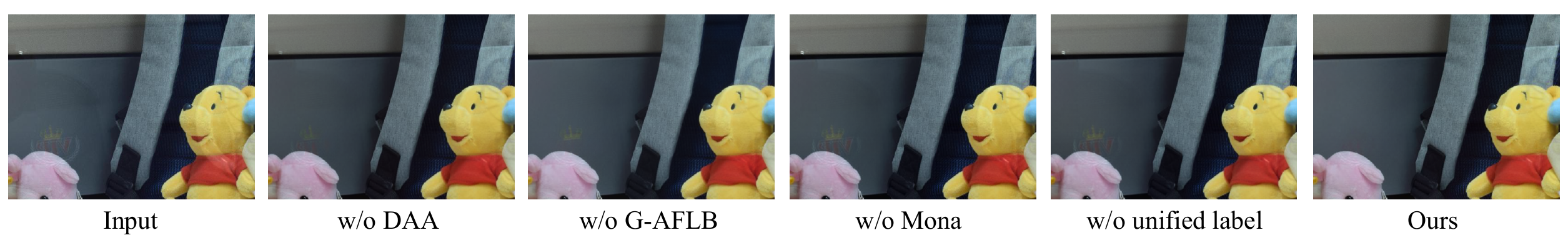}
	\caption{The visual comparison of different settings involved in the ablation study.}
	\label{fig:fig7}
\end{figure*}

\begin{figure}[t]
	\centering
	\includegraphics[width=0.99\linewidth]{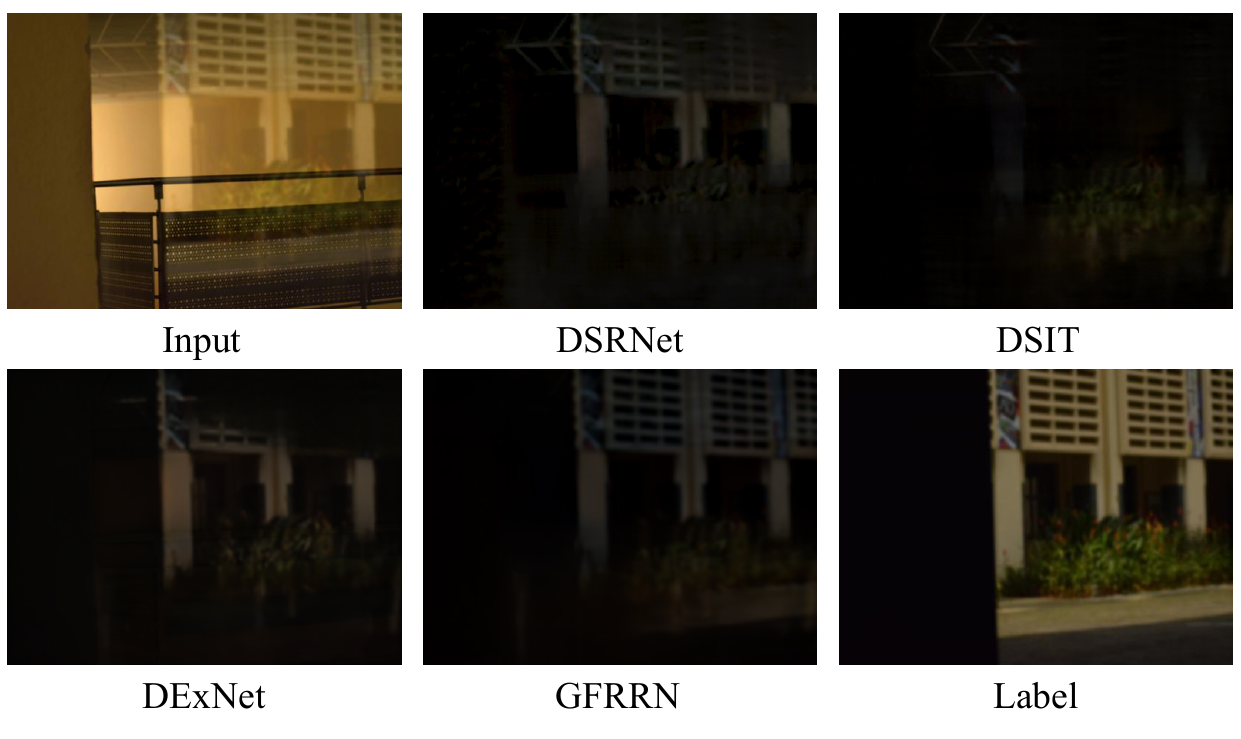}
	\caption{Visual comparison of the predicted reflection layers.}
	\label{fig:fig9}
\end{figure}

Fig.~\ref{fig:fig7} shows the visual results of different settings in Table~\ref{tab:tab2}.
Similar conclusion can be reached, with each component playing a positive role. 

\noindent \textbf{Mona-tuning.}
Rationally employing fine-tuning techniques can effectively mitigate the semantic gap between the pre-trained model and the reflection removal model. 
We conduct systematic ablation experiments on various fine-tuning techniques.
As shown in Table~\ref{tab:tab3}, the frozen model serves as the baseline, while full fine-tuning (FFT) and four parameter-efficient fine-tuning (PEFT) approaches are employed for comparison.

\begin{table}[h]
	\footnotesize
	\centering
	\caption{Ablation study on different fine-tuning techniques.}
	\label{tab:tab3}
	\begin{tabular}{l|cc}
		\toprule[1.2pt]
		Method & PSNR & SSIM \\
		\midrule\midrule
		(Baseline) Frozen & 26.70 & 0.920 \\
		\midrule
		FFT & 25.35 & 0.900 \\
		\midrule
		(PEFT) Bitfit & 26.86 & 0.923 \\
		(PEFT) Adaptformer & 27.32 & 0.925 \\
		(PEFT) LoRA & 26.51 & 0.920 \\
		\rowcolor{Gray}(PEFT) Mona & 27.33 & 0.929 \\
		\bottomrule[1.2pt]
	\end{tabular}
\end{table}

We observe that the FFT leads to a notable decline in performance. 
This is attributed to the huge parameter size of the pre-trained model, which does not align well with the scale of our dataset. 
In contrast, PEFT approaches yield favorable results, with the exception of LoRA. 
LoRA is suitable for NLP tasks, by adding parallel learnable matrices to multi-head attention weights. However, it proves to be not well-suited for visual tasks \cite{Yin2025CVPR-Mona}.
According to the performance, we choose Mona in our implementation.

\noindent \textbf{Unified label.}
According to the results in Table~\ref{tab:tab2}, unified label $(\mathbf{I} - \mathbf{T})_{\text{low}}$ is effective in our GFRRN.
We further conduct another experiment by employing $\mathbf{I}-\mathbf{T}$ as the unified reflection label.
As shown in Table~\ref{tab:tab4}, the performance drops from 27.33 dB to 26.61 dB. 
A possible explanation is that the $\mathbf{I}-\mathbf{T}$ contains certain high-frequency information originating from $\mathbf{T}$, which introduces ambiguity when attempting to separate reflection and transmission.
In addition, we also verify the effectiveness of our unified label on other architectures, such as DSIT \cite{Hu2024NeurIPS-DSIT} and DSRNet \cite{Hu2023ICCV-DSRNet}.

We further compare the reflection predictions against several dual-stream methods in Fig.~\ref{fig:fig9}.
We observe a favorable reflection in our GFRRN with clearer structures.
The buildings inside the reflection can be easily recognized.

\begin{table}[t]
	\footnotesize
	\centering
	\caption{Ablation study on proposed unified label with different architectures.}
	\label{tab:tab4}
	\begin{tabular}{c|c|cc}
		\toprule[1.2pt]
		Model & Setting & PSNR & SSIM \\
		\midrule\midrule
		\multirow{3}{*}{Ours}   & w/o unified label & 26.96     & 0.920     \\
		& w/ $\mathbf{I}-\mathbf{T}$ as unified label & 26.61 & 0.919 \\
		& \cellcolor{Gray}w/ unified label  & \cellcolor{Gray}27.33 & \cellcolor{Gray}0.929 \\
		\midrule
		\multirow{2}{*}{DSIT}   & w/o unified label & 26.50 & 0.919 \\
		& \cellcolor{Gray}w/ unified label  & \cellcolor{Gray}26.67 & \cellcolor{Gray}0.920 \\
		\midrule
		\multirow{2}{*}{DSRNet} & w/o unified label & 25.72 & 0.907 \\
		& \cellcolor{Gray}w/ unified label  & \cellcolor{Gray}26.17 & \cellcolor{Gray}0.915 \\
		\bottomrule[1.2pt]
	\end{tabular}
\end{table}

\noindent \textbf{G-AFLB and DAA.}
We also conducted ablation study on the designs of G-AFLB and DAA.
We fixed one of the components and conducted experiments on the other.
The experimental results are illustrated in Table~\ref{tab:tab5}.
It can be observed that our proposed G-AFLB achieves a 0.25 dB improvement in terms of PSNR compared to the AFLB.
This demonstrates the effectiveness of the Gaussian mask incorporated into our design.
As for DAA, it outperforms both W-MSA and agent attention by taking into account the varying degrees of importance among different windows.

\begin{figure}[t]
	\centering
	\includegraphics[width=0.99\linewidth]{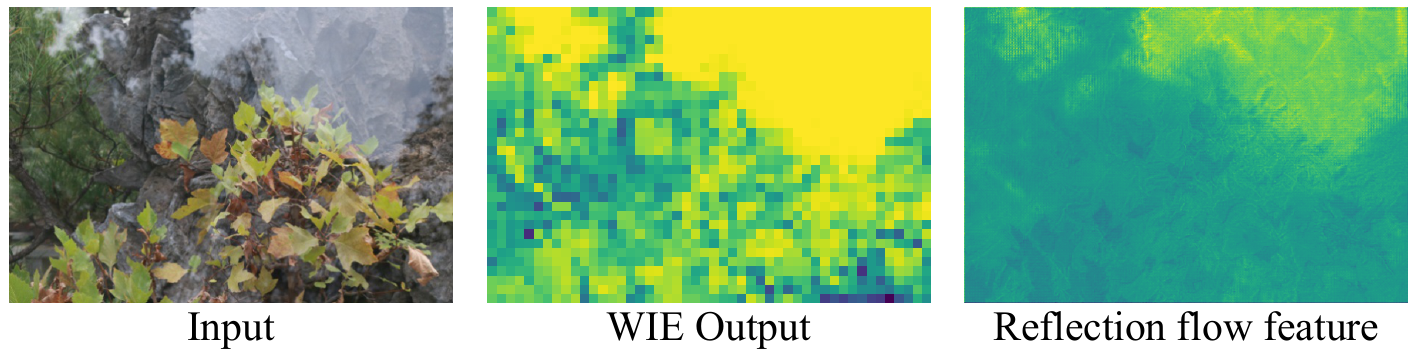}
	\caption{Given an input image with reflections, we visualize the importance weights map learned by WIE, and the feature of the reflection flow in our decoder.}
	\label{fig:fig8}
\end{figure}

Furthermore, we visualize the importance weights map learned by the window-based importance estimator (WIE) in Fig.~\ref{fig:fig8}.
After obtaining the output of WIE, the importance weight values are mapped back to the corresponding coordinate regions in the input image where the windows are located.
We also visualize the feature of the reflection flow in our decoder.
We choose the output feature of the decoder (before transferring back to three-channel RGB image) and do channel-wise average for displaying.
The learned importance weights map roughly indicates the regions where reflections are present.

\begin{table}[t]
	\footnotesize
	\centering
	\caption{Ablation study on different settings in our decoder.}
	\label{tab:tab5}
	\begin{tabular}{cc|cc}
		\toprule[1.2pt]
		\multicolumn{2}{c|}{Setting}                 & PSNR & SSIM \\ 
		\midrule\midrule
		\multirow{2}{*}{DAA \quad+}    & AFLB (hard mask)            & 27.08    & 0.926    \\
		& \cellcolor{Gray}G-AFLB (Gaussion mask)          & \cellcolor{Gray}27.33    & \cellcolor{Gray}0.929    \\
		\midrule
		\multirow{3}{*}{G-AFLB \quad+}  & W-MSA  & 26.91  & 0.919  \\
		& Agent Attention & 27.04    & 0.924 \\
		& \cellcolor{Gray}DAA             & \cellcolor{Gray}27.33    & \cellcolor{Gray}0.929    \\ 
		\bottomrule[1.2pt]
	\end{tabular}
\end{table}

%

%
\section{Conclusion}
\label{sec:conclusion}


In conclusion, this paper presents the GFRRN, a novel approach designed to address the SIRR task. 
By integrating a parameter-efficient fine-tuning strategy (\ie, Mona-tuning), our method effectively bridges the semantic gap between pre-trained and restoration models. 
Additionally, the introduction of a unified label generator ensures consistent supervision across both synthetic and real-world datasets, thereby mitigating the training data gap. 
The proposed 
G-AFLB and DAA further enhance the model's capability to adaptively learn and fuse frequency priors while dynamically modeling inter- and intra-window significance levels.

{
    \small
    \bibliographystyle{ieeenat_fullname}
    \bibliography{main}
}

\clearpage
\setcounter{page}{1}
\maketitlesupplementary

\appendix


\section{The details of Mona Layer}
\begin{figure}[t]
	\centering
	\includegraphics[width=1\linewidth]{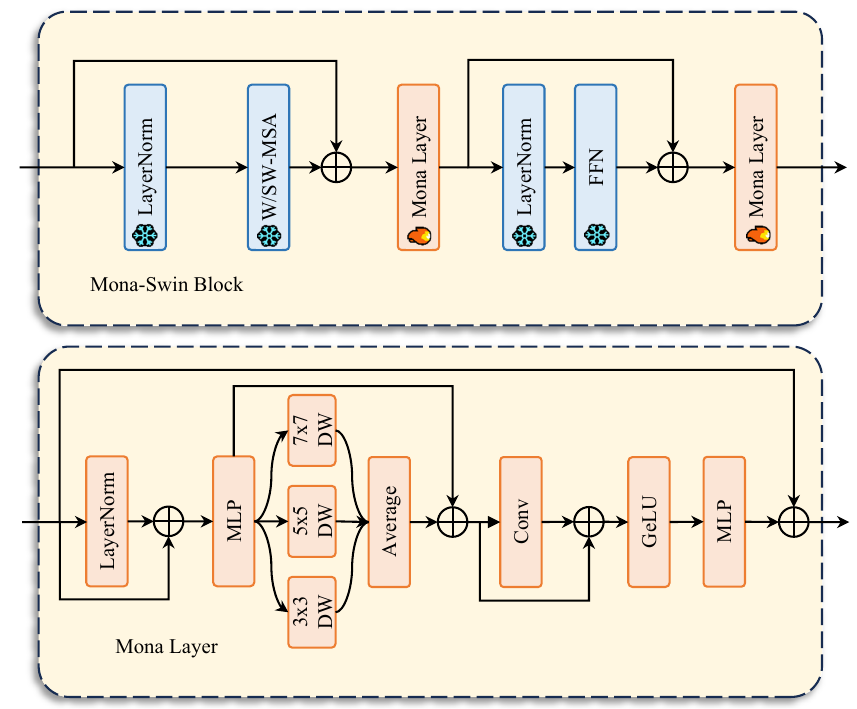}
	\caption{Top: The detail of Mona-Swin Block. Bottom: The detail of Mona Layer.}
	\label{fig:fig10}
\end{figure}

Fig.~\ref{fig:fig10} illustrates the position of the Mona layer \cite{Yin2025CVPR-Mona} within the SwinBlock \cite{Liu2021ICCV-Swin} and presents its detailed structure. 
The Mona layer is inserted after the Attention and Feed-Forward Network to fine-tune their outputs. 
The core of the Mona layer is a set of visual filters composed of multi-scale depthwise convolutions ($3\times 3$, $5\times 5$, $7\times 7$) and a pointwise convolution ($1\times 1$), which collectively capture multi-scale visual information.

\section{The overall structure of decoder}
Alg.~\ref{alg:decoder_layer_detailed} outlines the overall workflow of a certain level in the decoder. Each level consists of the G-AFLB and the Dual-stream Dynamic Interaction Block (DDIB). 
The DDIB is a dual-stream Transformer block that incorporates DAA and Layer-wise DAA (\ie, LDAA). 

\begin{algorithm}
\caption{Structure of Certain Level of Decoder}
\label{alg:decoder_layer_detailed}
\begin{algorithmic}[1]
\REQUIRE{Input features ${F}_\textbf{T}^{0}$, ${F}_\textbf{R}^{0}$ and degraded image \textbf{I}}
\ENSURE{Output features ${F}_\textbf{T}^{K}$ and ${F}_\textbf{R}^{K}$}

\STATE \textbf{Step1: Apply G-AFLB}
\STATE \quad ${F}_\textbf{T}^{1} = \text{G-AFLB}({F}_\textbf{T}^{0}, \textbf{I})$
\STATE \quad ${F}_\textbf{R}^{1} = \text{G-AFLB}({F}_\textbf{R}^{0}, \textbf{I})$

\STATE \textbf{Step2: Apply K DDIB}
\STATE \quad ${F}_\textbf{T}^{{current}} = {F}_\textbf{T}^{1}$, ${F}_\textbf{R}^{{current}} = {F}_\textbf{R}^{1}$
\FOR{$i = 1$ \TO $\text{K}$}
\STATE \quad \textbf{LayerNorm:} 
\STATE \quad \quad ${F}_\textbf{T}^{LN} = {LN}({F}_\textbf{T}^{{current}})$
\STATE \quad \quad ${F}_\textbf{R}^{LN} = {LN}({F}_\textbf{R}^{{current}})$
\STATE \quad \textbf{Combine tokens:}
\STATE \quad \quad ${X}_{0}^{LN} = \text{Concat}([{F}_\textbf{T}^{LN}, {F}_\textbf{R}^{LN}], {dim=0})$
\STATE \quad \quad ${X}_{1}^{LN} = \text{Concat}([{F}_\textbf{T}^{LN}, {F}_\textbf{R}^{LN}], {dim=1})$
\STATE \quad \textbf{Apply attention:} 
\STATE \quad \quad ${X}_{0}^{SA} = {DAA}({X}_{0}^{LN})$
\STATE \quad \quad ${X}_{1}^{CA} = {LDAA}({X}_{1}^{LN})$
\STATE \quad \textbf{Split back:}
\STATE \quad \quad ${F}_\textbf{T}^{SA}, {F}_\textbf{R}^{SA} = \text{Split}({X}_{0}^{SA}, {dim=0})$
\STATE \quad \quad ${F}_\textbf{T}^{CA}, {F}_\textbf{R}^{CA} = \text{Split}({X}_{1}^{CA}, {dim=1})$
\STATE \quad \textbf{Combine the dual-attention results:} 
\STATE \quad \quad ${F}_\textbf{T}^{DA} = {F}_\textbf{T}^{{current}} + {F}_\textbf{T}^{SA} +  {F}_\textbf{T}^{CA}$
\STATE \quad \quad ${F}_\textbf{R}^{DA} = {F}_\textbf{R}^{{current}} + {F}_\textbf{R}^{SA} +  {F}_\textbf{R}^{CA}$
\STATE \quad \textbf{Apply FFN:} 
\STATE \quad \quad ${F}_\textbf{T}^{LN'} = {LN}({F}_\textbf{T}^{{DA}})$
\STATE \quad \quad ${F}_\textbf{R}^{LN'} = {LN}({F}_\textbf{R}^{{DA}})$
\STATE \quad \quad ${F}_\textbf{T}^{FFN}, {F}_\textbf{R}^{FFN} = \text{DSLPBlock}({F}_\textbf{T}^{LN'}, {F}_\textbf{R}^{LN'})$
\STATE \quad \textbf{Output:}
\STATE \quad \quad ${F}_\textbf{T}^{{current}} = {F}_\textbf{T}^{DA} + {F}_\textbf{T}^{FFN}$
\STATE \quad \quad ${F}_\textbf{R}^{{current}} = {F}_\textbf{R}^{DA} + {F}_\textbf{R}^{FFN}$
\ENDFOR

\STATE \textbf{Output:} ${F}_\textbf{T}^{K} = {F}_\textbf{T}^{{current}}$, ${F}_\textbf{R}^{K} = {F}_\textbf{R}^{{current}}$
\end{algorithmic}
\end{algorithm}

\section{The details of G-AFLB}
\begin{figure}[htb]
	\centering
	\includegraphics[width=0.99\linewidth]{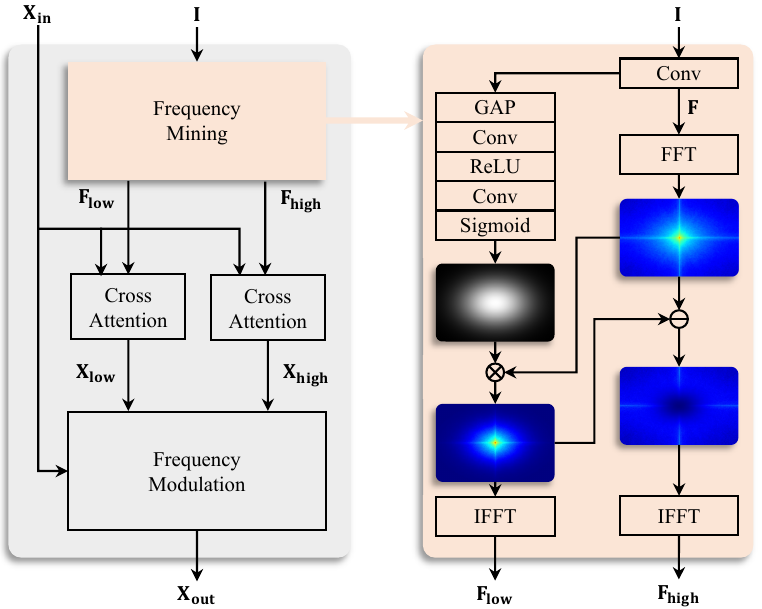}
	\caption{Overall structure of G-AFLB and details of FMiM.}
	\label{fig:fig11}
\end{figure}

The Adaptive Frequency Learning Block (AFLB) \cite{Cui2025ICLR-adaIR} can be divided into Frequency Mining Module (FMiM) and Frequency Modulation Module (FMoM). Its function is to modulate the feature maps using the frequency information of the input image \textbf{I}. Specifically, the FMiM adaptively separates the high-low frequency components of the enhanced input image, while the FMoM modulates the input feature map with the separated frequency components, enabling the feature map to explicitly carry frequency information. G-AFLB's key is the FMiM, which separates high-low frequency through Gaussian low-pass filter. The details are shown in Fig.~\ref{fig:fig11}. This process can be formally represented as follows:

\begin{equation}
\begin{gathered}
{F}_{{low}}, {F}_{{high}}=\text{FMiM}(\textbf{I}), \\
{X}_{{low}}=\text{CrossAttention}\left({F}_{{low}}, {X}_{in}\right), \\
{X}_{{high}}=\text{CrossAttention}\left({F}_{{high}}, {X}_{in}\right), \\
{X}_{{out}}=\text{FMoM}\left({F}_{{low}}, {F}_{{high}}, {X}_{in}\right).
\end{gathered}
\end{equation}

The original FMiM used rectangular masks in the frequency domain to separate the high-low frequency information of the input image \textbf{I}. However, this can lead to ringing artifacts in the corresponding spatial domain image. The specific derivation is provided below:

Let $(x, y)$ denote spatial coordinates and $(\omega_{x},\omega_{y})$ denote the corresponding frequency domain coordinates. Let $f(x, y)$ be a function in the spatial domain, and $F(\omega_x,\omega_y)$ be its frequency domain response. The formulas for the Fourier transform and its inverse are:

\begin{equation}
\begin{gathered}
F\left(\omega_x,\omega_y\right)=\iint_{\mathbb{R}^2} f(x, y) e^{-i\left(\omega_x x+\omega_y y\right)}  d x  d y, \\
f(x, y)=\frac{1}{(2\pi)^2}\iint_{\mathbb{R}^2} F\left(\omega_x,\omega_y\right) e^{i\left(\omega_x x+\omega_y y\right)}  d\omega_x  d\omega_y.
\end{gathered}
\end{equation}

According to the convolution theorem, multiplication in the frequency domain is equivalent to convolution in the spatial domain. Therefore, multiplying by a mask in the frequency domain is equivalent to convolving with the spatial domain response of that mask in the spatial domain:

\begin{equation}
\mathcal{F}\{f * g\}=F\left(\omega_x,\omega_y\right)\cdot G\left(\omega_x,\omega_y\right).
\end{equation}

A frequency domain mask is essentially a low-pass filter. The frequency domain expression for an ideal rectangular low-pass filter is:

\begin{equation}
H_{Rec}\left(\omega_x,\omega_y\right)=
\begin{cases}
1, & \left|\omega_x\right|\leq\omega_{c_x} \text{ and } \left|\omega_y\right|\leq\omega_{c_y} \\
0, & \text{ else }
\end{cases},
\end{equation}

Substituting $H_{Rec}$ into the inverse Fourier transform formula yields its spatial domain impulse response:

\begin{align*}
h_{Rec}(x, y) &= \frac{1}{(2\pi)^2} \iint\limits_{\left|\omega_x\right| \leq \omega_{c_x}, \left|\omega_y\right| \leq \omega_{c_y}} e^{i\left(\omega_x x+\omega_y y\right)}  d\omega_x  d\omega_y \\
&= \frac{1}{(2\pi)^2} \left(\int_{-\omega_{c_x}}^{\omega_{c_x}} e^{i\omega_x x}  d\omega_x\right)
\left(\int_{-\omega_{c_y}}^{\omega_{c_y}} e^{i\omega_{y} y}  d\omega_y\right)
\end{align*}
\begin{equation}
\hspace{-2.5em}=\frac{1}{\pi^{2}}\frac{\sin\left(\omega_{c_{x}}x\right)}{x}\cdot\frac{\sin\left(\omega_{c_{y}}y\right)}{y}.
\end{equation}

This result shows that the spatial domain response of the ideal rectangular low-pass filter is a two-dimensional Sinc function. This function is not single-peaked and oscillates between positive and negative values. When convolved with image edges, this oscillatory response causes oscillations in the image, manifesting as the ringing artifact.

The frequency domain representation of a Gaussian low-pass filter is:

\begin{equation}
H_{G}\left(\omega_{x},\omega_{y}\right)=e^{-\frac{1}{2}\left(\frac{\omega_{x}^{2}}{\sigma_{x}^{2}}+\frac{\omega_{y}^{2}}{\sigma_{y}^{2}}\right)},\quad \sigma_{x}>0,\sigma_{y}>0,
\end{equation}

Substituting this into the inverse transform formula:

$$
h_G(x,y)=\frac{1}{(2\pi)^2}\iint_{\mathbb{R}^2} e^{-\frac{1}{2}\left(\frac{\omega_x^2}{\sigma_x^2}+\frac{\omega_y^2}{\sigma_y^2}\right)} e^{i\left(\omega_x x+\omega_y y\right)} d\omega_x d\omega_y
$$

\begin{equation}
=\frac{1}{(2\pi)^2}\left(\int_{\mathbb{R}} e^{-\frac{\omega_x^2}{2\sigma_x^2}} e^{i\omega_x x} d\omega_x\right)\left(\int_{\mathbb{R}} e^{-\frac{\omega_y^2}{2\sigma_y^2}} e^{i\omega_y y} d\omega_y\right),
\end{equation}

Using the Gaussian integral formula:

\begin{equation}
\int_{-\infty}^{\infty} e^{-a\omega^2} e^{j\omega x}  d\omega=\sqrt{\frac{\pi}{a}} e^{-\frac{x^2}{4a}},
\end{equation}

For the x-direction integral, let $a=\frac{1}{2\sigma_{x}^{2}}$. For the y-direction integral, let $a=\frac{1}{2\sigma_{y}^{2}}$. This yields:

\begin{equation}
\int_{\mathbb{R}} e^{-\frac{\omega_{x}^{2}}{2\sigma_{x}^{2}}}e^{j\omega_{x} x}  d\omega_{x}=\sqrt{2\pi\sigma_{x}^{2}} e^{-\frac{\sigma_{x}^{2} x^{2}}{2}},
\end{equation}
\begin{equation}
\int_{\mathbb{R}} e^{-\frac{\omega_{y}^{2}}{2\sigma_{y}^{2}}}e^{j\omega_{y} y}  d\omega_{y}=\sqrt{2\pi\sigma_{y}^{2}} e^{-\frac{\sigma_{y}^{2} y^{2}}{2}},
\end{equation}

Combining these results:

\begin{equation}
\begin{gathered}
h_{G}(x, y)=\frac{1}{(2\pi)^2}\cdot\sqrt{2\pi\sigma_x^2} e^{-\frac{\sigma_x^2 x^2}{2}}\cdot\sqrt{2\pi\sigma_y^2} e^{-\frac{\sigma_y^2 y^2}{2}} \\
=\frac{\sigma_x\sigma_y}{2\pi} e^{-\frac{1}{2}\left(\sigma_x^2 x^2+\sigma_y^2 y^2\right)}.
\end{gathered}
\end{equation}

Therefore, the spatial domain response of the Gaussian frequency domain filter is still a two-dimensional Gaussian function. The Gaussian function is single-peaked and always positive. Convolving this response with a spatial image results in blurring without introducing ringing artifacts.


\section{The details of DAA and LDAA}
We propose the DAA mechanism, a novel self-attention variant that accounts for content differences across windows. 
The overall procedure is outlined in Alg.~\ref{alg:dynamic_agent_attention}, where DWC denotes depthwise convolution.
\begin{figure}[htb]
	\centering
	\includegraphics[width=1\linewidth]{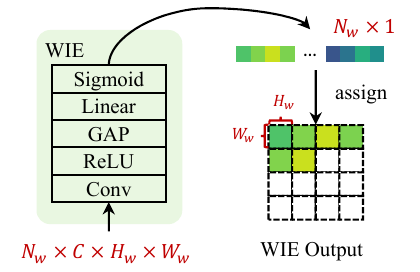}
	\caption{The structure of our proposed WIE.}
	\label{fig:fig13}
\end{figure}

A lightweight WIE module is introduced to predict the difference in reflection intensity between windows. This design can be generalized to all window-based attention mechanisms, with detailed structure shown in Fig.~\ref{fig:fig13}. The Q vector is reshaped to dimensions $(N_w, C, H_w, W_w)$ and fed into the WIE module, where $N_w$ represents the number of windows, and $H_w$ and $W_w$ denote the height and width of each window, respectively. The WIE module outputs a vector of shape $(N_w, 1)$, assigning a weight to each window. For visualization purposes, the resulting weights are remapped to the spatial dimensions $(1, 1, H, W)$, where $H$ and $W$ are the height and width of the feature map.

We further extend the DAA to a method termed LDAA,  which enables dual-stream interaction between features of the transmissive layer and the reflective layer. The complete procedure is described in Alg.~\ref{alg:layered_dynamic_agent_attention}.

\section{New benchmark and additional visual comparisons}
We have additionally captured a new testing dataset named GF40. 
It consists of 40 image pairs. 
Each pair includes a superimposed image $\textbf{I}$ and its transmission layer image $\textbf{T}$. 

\begin{figure}[htb]
	\centering
	\includegraphics[width=1\linewidth]{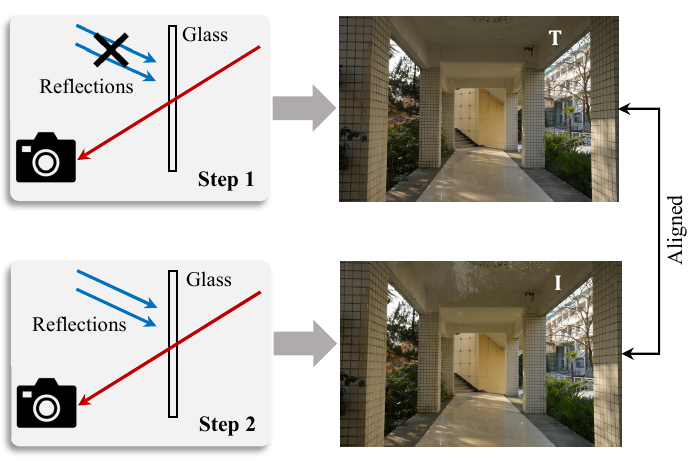}
	\caption{The capturing process of GF40.}
	\label{fig:fig14}
\end{figure}

The capturing process is illustrated in Fig. ~\ref{fig:fig14}. 
Following \cite{Zhu2024CVPR-RRW}, we first capturing $\textbf{T}$ by blocking the light source on the same side as the camera, and then capturing $\textbf{I}$ without any obstruction. 
This approach can prevent pixel misalignment caused by glass refraction.

As a supplement to existing testing datasets, GF40 is captured under the following conditions:
\begin{itemize}
  \item Camera: Sony ILCE-7RM4A
  \item Glass thickness: 3 mm
  \item Shooting scenes: Indoor and outdoor
  \item Camera aperture: f/2.0--f/16
  \item Contains instances of strong reflections
\end{itemize}

\begin{figure*}[!p]
	\centering
	\includegraphics[width=1\linewidth]{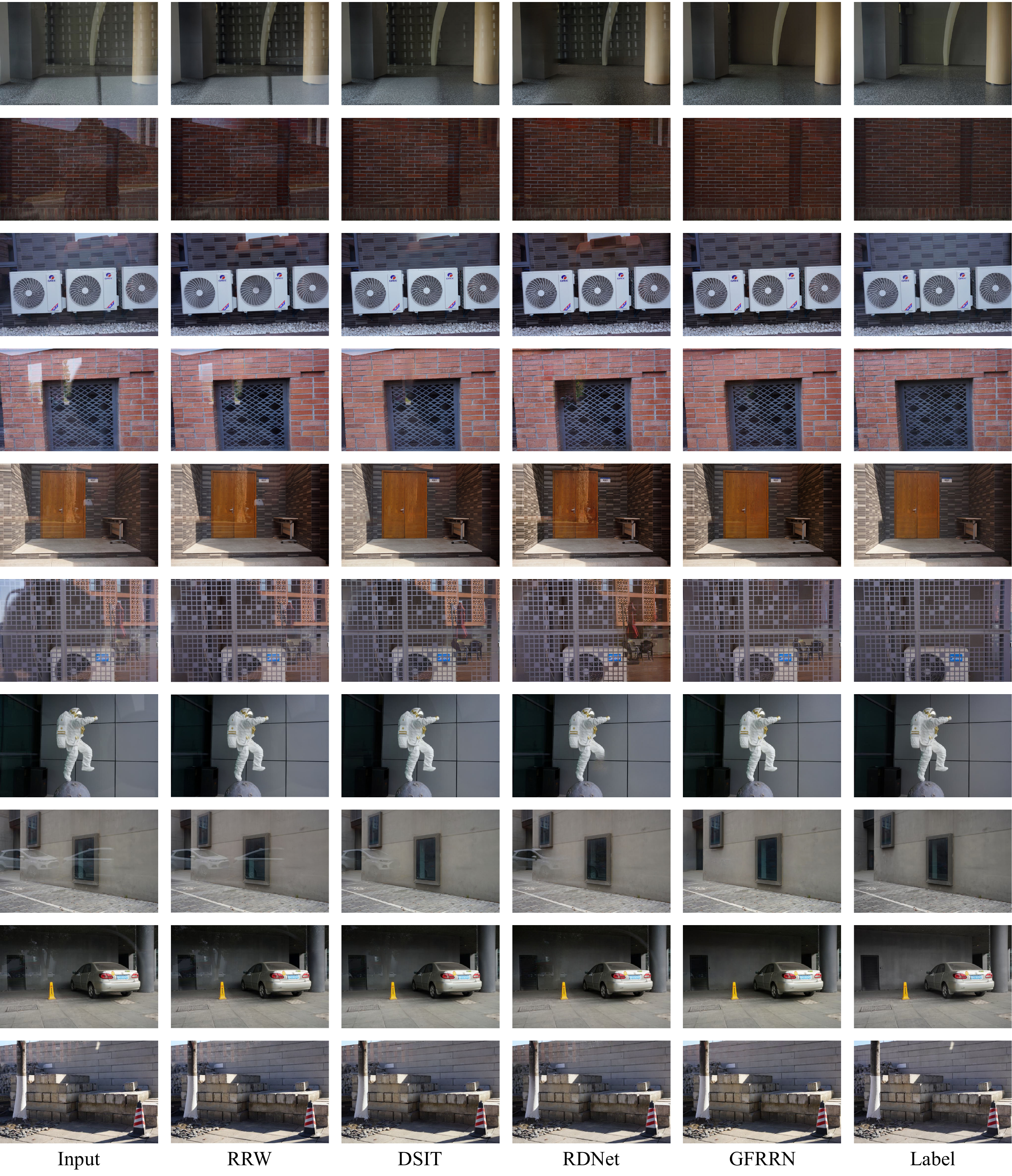}
	\caption{The visual results of RRW, DSIT, RDNet and our GFRRN on samples from GF40.}
	\label{fig:fig15}
\end{figure*}

We test and compare the state-of-the-art methods on the GF40 dataset. 
As shown in Tab.~\ref{tab:tab6}, the quantitative comparisons are provided, while qualitative results are visualized in Fig.~\ref{fig:fig15}.

\begin{table}[htb]
	\footnotesize
	\caption{Performance comparison of our GFRRN and current state-of-the-art methods on GF40. We use the pre-trained models they provided for evaluation. The best results are displayed in \textbf{bold}, while the second-best are \underline{underlined}.}
	\label{tab:tab6}
	\centering
	\begin{tabular}{l|l|cc}
		\toprule[1.2pt]
		Methods & Venues& PSNR & SSIM \\
		\midrule\midrule
        DSRNet \cite{Hu2023ICCV-DSRNet} & ICCV'23& 23.79 & 0.860 \\
        DURRNet \cite{Huang2024ICASSP-DURRnet}& ICASSP'24& 22.43 & 0.815\\
		RRW \cite{Zhu2024CVPR-RRW}& CVPR'24& 23.65 & 0.863 \\
		DSIT \cite{Hu2024NeurIPS-DSIT}&  NeurIPS'24& \underline{24.98} & \underline{0.868} \\
        DExNet \cite{Huang2025TPAMI-DExNet}& TPAMI'25& 23.16 & 0.845 \\
		RDNet \cite{Zhao2025CVPR-RDNet}& CVPR'25&24.35 & 0.856 \\
		\midrule
		\rowcolor{Gray}GFRRN & -& \textbf{25.95} & \textbf{0.876} \\
		\bottomrule[1.2pt]
	\end{tabular}
\end{table}


\begin{algorithm}
\caption{Dynamic Agent Attention}
\label{alg:dynamic_agent_attention}
\begin{algorithmic}[1]
\REQUIRE{Input tensor ${x}_{{in}} \in \mathbb{R}^{2{N}_{{w}} \times {H}_{{w}}{W}_{{w}} \times {C}}$}
\ENSURE{Output tensor ${x}_{{out}} \in \mathbb{R}^{2{N}_{{w}} \times {H}_{{w}}{W}_{{w}} \times {C}}$}
\STATE \textbf{Step1: Compute Q, K, V}
\STATE \quad ${Q},{K},{V} = \text{Linear}({x}_{{in}})$ 
\STATE \textbf{Step2: Agent Generation}
\STATE \quad ${A} = \text{AgentGenerate}({Q})$
\STATE \textbf{Step3: Window Importance Estimation}
\STATE \quad ${score} = \text{WIE}({Q})$
\STATE \quad ${A}_{{w}} = {A} \cdot {score}$
\STATE \textbf{Step4: Agent Aggregation}
\STATE \quad ${V}_{{A}} = \text{softmax}({A}_{{w}} \cdot {K}^{\top} + {bias}) \cdot {V}$
\STATE \textbf{Step5: Agent Broadcast}
\STATE \quad ${F}^{{attn}} = \text{softmax}({Q} \cdot {A}_{{w}}^{\top} + {bias}) \cdot {V}_{{A}}$
\STATE \textbf{Step6: Feature Enhancement}
\STATE \quad ${F}^{{dwc}} = \text{DWC}({V})$
\STATE \textbf{Step7: Output Projection}
\STATE \quad ${x}_{{out}} = \text{Linear}({F}^{{attn}} + {F}^{{dwc}})$
\STATE \textbf{Output:} ${x}_{{out}}$
\end{algorithmic}
\end{algorithm}

\begin{algorithm}
\caption{Layer-wise Dynamic Agent Attention}
\label{alg:layered_dynamic_agent_attention}
\begin{algorithmic}[1]
\REQUIRE{Input tensor ${x}_{{in}} \in \mathbb{R}^{{N}_{{w}} \times 2{H}_{{w}}{W}_{{w}} \times {C}}$}
\ENSURE{Output tensors ${x}_{{out}} \in \mathbb{R}^{{N}_{{w}} \times 2{H}_{{w}}{W}_{{w}} \times {C}}$}
\STATE \textbf{Step1: Compute Q, K, V}
\STATE \quad ${Q},{K},{V} = \text{Linear}({x}_{{in}})$
\STATE \textbf{Step2: Layer-wise Agent Generation}
\STATE \quad ${Q}_{\textbf{T}},{Q}_{\textbf{R}} = \text{LayerSeparate}({Q})$
\STATE \quad ${A}_{\textbf{T}} = \text{AgentGenerate}({Q}_{\textbf{T}})$
\STATE \quad ${A}_{\textbf{R}} = \text{AgentGenerate}({Q}_{\textbf{R}})$
\STATE \quad ${A} = \text{LayerCombine}({A}_{\textbf{T}}, {A}_{\textbf{R}})$
\STATE \textbf{Step3: Layer-wise Window Importance Estimation}
\STATE \quad ${score}_{\textbf{T}} = \text{WIE}({Q}_{\textbf{T}}), {score}_{\textbf{R}} = \text{WIE}({Q}_{\textbf{R}})$
\STATE \quad ${score} = \text{Average}({score}_{\textbf{T}},{score}_{\textbf{R}})$
\STATE \quad ${A}_{{w}} = {A} \cdot {score}$
\STATE \textbf{Step4: Layer-wise Agent Aggregation}
\STATE \quad ${V}_{{A}} = \text{softmax}({A}_{{w}} {K}^\top + {bias}_{{layered}}) {V}$
\STATE \textbf{Step5: Layer-wise Agent Broadcast}
\STATE \quad ${F}^{{attn}} = \text{softmax}({Q} {A}_{{w}}^\top + {bias}_{{layered}}) {V}_{{A}}$
\STATE \textbf{Step6: Layer-wise Feature Enhancement}
\STATE \quad ${V}_{\textbf{T}},{V}_{\textbf{R}} = \text{LayerSeparate}({V})$
\STATE \quad ${F}_{\textbf{T}}^{{dwc}} = \text{DWC}({V}_{\textbf{T}})$
\STATE \quad ${F}_{\textbf{R}}^{{dwc}} = \text{DWC}({V}_{\textbf{R}})$
\STATE \quad ${F}^{{dwc}} = \text{LayerCombine}({F}_{\textbf{R}}^{{dwc}}, {F}_{\textbf{T}}^{{dwc}})$
\STATE \textbf{Step7: Output Projection}
\STATE \quad ${x}_{{out}} = \text{Linear}({F}^{{attn}} + {F}^{{dwc}})$
\STATE \textbf{Output:} ${x}_{{out}}$
\end{algorithmic}
\end{algorithm}


\end{document}